\documentclass[11pt, a4paper]{article}
\pdfoutput=1
\usepackage{times}
\usepackage{graphicx}
\usepackage{amsmath}
\usepackage{amssymb}
\usepackage{subfigure}
\usepackage{textcomp}
\usepackage{soul}
\usepackage{multirow}
\usepackage{bm}
\usepackage{algorithm,algorithmicx}
\usepackage[noend]{algpseudocode}
\usepackage{tikz}
\definecolor{newblue}{rgb}{0.2,0.2,0.6} 

\usepackage[margin=2cm]{geometry}
\usepackage{tikz,subfigure,sectsty}

\newcommand{\vecx}{\mathbf{x}}

\newcommand{\vecc}{\mathbf{c}}

\DeclareMathOperator*{\argmin}{arg\,min}
\DeclareMathOperator{\Tr}{Tr}

\newcommand{\R}{\mathbb{R}}

\definecolor{ballblue}{rgb}{0.13, 0.67, 0.8}
	\definecolor{newbrown}{rgb}{0.65, 0.16, 0.16}	
\definecolor{cblue}{rgb}{0.64, 0.76, 0.68}
	\definecolor{cerise}{rgb}{0.87, 0.19, 0.39}
		\definecolor{jasper}{rgb}{0.84, 0.23, 0.24}
	\definecolor{mblue}{rgb}{0.15, 0.38, 0.61}

\begin{document}

\title{Temporal Human Action Segmentation via Dynamic Clustering}

\author{Yan Zhang\\
Institute of Neural Information Processing, Ulm University, Germany\\
{\tt\small yan.zhang@uni-ulm.de}
\and
He Sun\\
School of Informatics, The University of Edinburgh, UK\\
{\tt\small h.sun@ed.ac.uk}
\and
Siyu Tang\\
Department of Perceiving Systems, Max Planck Institute for Intelligent Systems, Germany\\
The University of T\"{u}bingen, Germany\\
{\tt\small 	stang@tuebingen.mpg.de}
\and
Heiko Neumann\\
Institute of Neural Information Processing, Ulm University, Germany\\
{\tt\small  heiko.neumann@uni-ulm.de}
}

\maketitle

\begin{center}
\begin{minipage}[c]{0.75\linewidth}
    \begin{abstract} 

We present an effective dynamic clustering algorithm for the task of temporal human action segmentation, which has comprehensive applications such as robotics, motion analysis, and patient monitoring. Our proposed algorithm is unsupervised, fast, generic to process various types of features, and applicable in both the online and offline settings.  We perform extensive experiments of  processing data streams, and show that our algorithm achieves the state-of-the-art results for both online and offline settings.
\end{abstract}
\end{minipage}
\end{center}

\section{Introduction}

Automatic analysing human behaviours in a video stream is an important step for building an intelligent system that can perceive and interact with humans as humans do. It closely relates to many computer vision tasks, such as video summarisation, segmentation of human motion sequences~\cite{ACA}, and detection of unusual activities~\cite{ZSV04}, which have received a lot of attention in recent years.

 Most of previous algorithms for analysing human behaviours are based on supervised learning, in which a large amount of training sets are required and human behaviours in each training set need to be labelled manually. In addition,
most supervised learning approaches focus on the offline setting, producing the result after processing an entire video. While the performance of such supervised approach also relies on the quality of annotation and the irregularity in the periodicity of human actions, for many important applications like health-care and surveillance the training sets are usually not publicly accessible for data privacy reasons. Thus, developing an unsupervised and data-driven approach to analysing human behaviours  is of great interest.

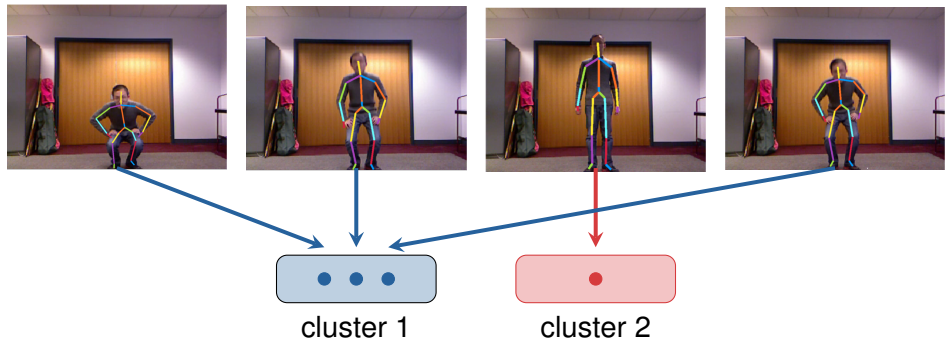
\begin{figure}[htb]
\begin{center}
\begin{tikzpicture}[xscale=1.05,yscale=1.05,
rknoten/.style={fill=jasper,color=jasper,draw=jasper,circle,scale=0.45},edge/.style={black, thick},
gknoten/.style={fill=newgreen,color=newgreen,draw=newgreen,circle,scale=0.45},edge/.style={black, thick},
bknoten/.style={fill=mblue,color=mblue,draw=mblue,circle,scale=0.45},edge/.style={black, thick},
redge/.style={draw=red!70,thick},
gedge/.style={draw=newgreen,thick},
bedge/.style={draw=newblue,thick},
edge/.style={black, thick}]


\filldraw[fill=mblue!30, rounded corners=4pt] (2,-0.7) rectangle (4, -0.1);

\filldraw[draw=jasper, fill=jasper!30, rounded corners=4pt] (5,-0.7) rectangle (7,-0.1);

\node at (0,2) { \includegraphics[width=0.17\linewidth]{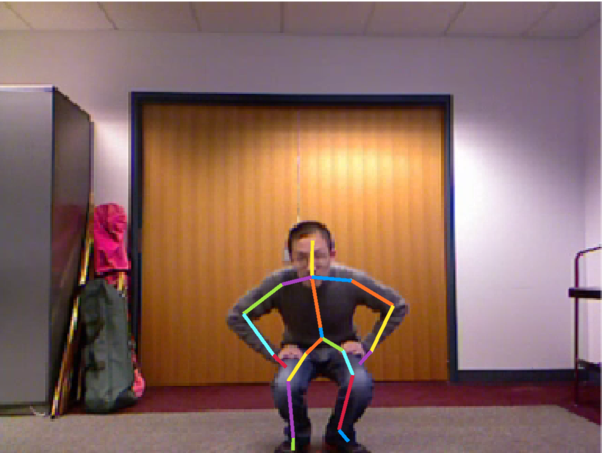}};

\node[bknoten] (1) at (2.6,-0.4){};
\draw[-stealth, line width=1.5pt, color=mblue] (0,1.0) -- (2.6,-0);

\node at (3,2) { \includegraphics[width=0.17\linewidth]{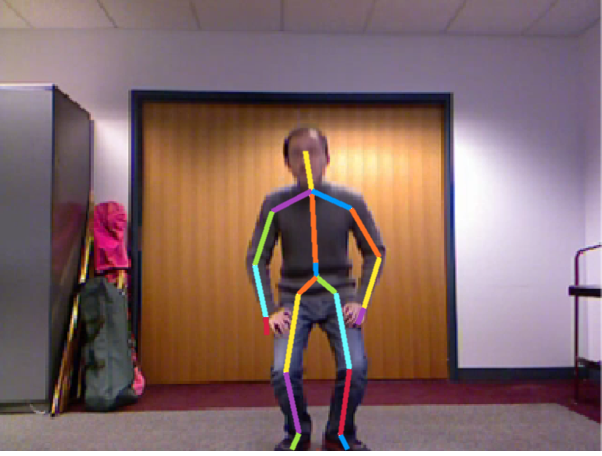}};

\node[bknoten] (1) at (3,-0.4){};
\draw[-stealth, line width=1.5pt, color=mblue] (3,1.0) -- (3,0);

\node[bknoten] (1) at (3.4,-0.4){};

\node at (6,2) { \includegraphics[width=0.17\linewidth]{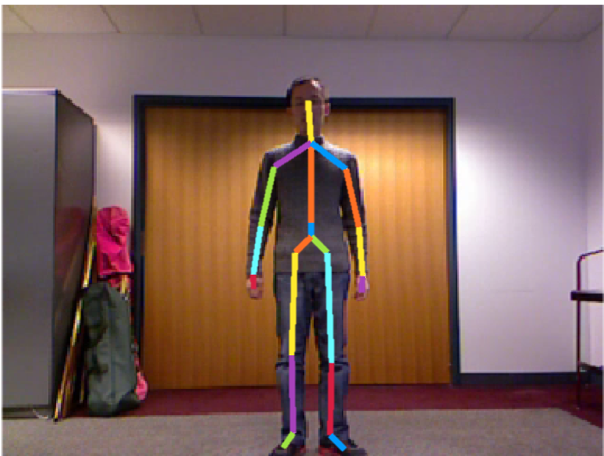}};

\node[rknoten] (1) at (6,-0.4){};
\draw[-stealth, line width=1.5pt, color=jasper] (6,1) -- (6,0);

\node at (9,2) { \includegraphics[width=0.17\linewidth]{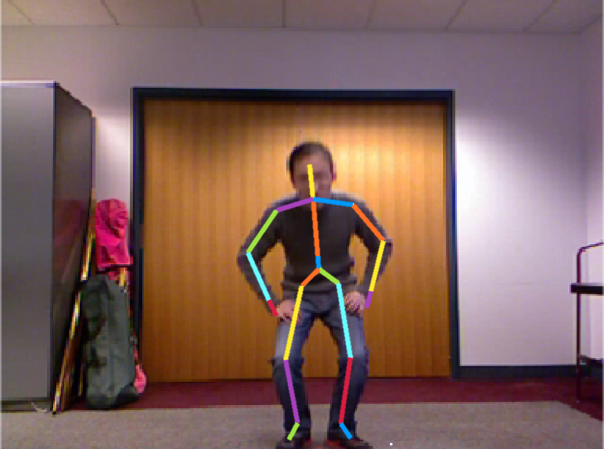}};

\draw[-stealth, line width=1.5pt, color=mblue] (9,1) -- (3.4,-0);

\node at (6,-1) {\textsf{cluster 2}};
\node at (3,-1) {\textsf{cluster 1}};

\end{tikzpicture}
\end{center}\vspace{-0.5cm}
   \caption{Illustration of our  approach. Our algorithm updates the clustering structure dynamically, and performs the dynamic segmentation task based on the transition point from which two consecutive feature vectors belong to different clusters. \label{fig:overview}}

\end{figure}

In this work we propose a fast and  dynamic clustering algorithm to process a generic multi-dimensional time series. 
As the output of our algorithm, the clusters are formed in a data-driven fashion and evolve dynamically over time, see Figure~\ref{fig:overview} for illustration. More specifically, our algorithm consists of two steps: (1) In the initialisation step, we utilise a robust spectral clustering method~\cite{njw,PSZ} to learn the centers and the covariances of the initial patterns in an unsupervised fashion. These basic statistics, including the centers and covariances of the clusters, are associated with the spatial concentration/separation properties of the same/different human actions of feature vectors, and are used to predict new human actions; (2) in the online evolution step, the algorithm either treats every arriving feature vector as one of the existing actions or recognises it as a new action, and further updates the center and covariance information of  an existing or new cluster. Due to the ability of effectively extracting statistical information of feature vectors, our algorithm is able to process video streams from generic multi-dimensional time series and different kinds of features.

We perform a number of  experiments for the proposed algorithm, and compare it with the state-of-the-art.
 For instance,  
when applying the  \textsf{JointLocation}  features to the {\sf CMUMAD} dataset~\cite{huang2014sequential}, our method achieves $0.82/0.86$ precision/recall value within $0.1$ seconds in the offline setting. The classical offline Aligned Cluster Analysis (\textsf{ACA}) algorithm~\cite{ACA} achieves $0.55/0.68$ precision/recall value, but its runtime is $2,200$ times slower than ours. 

We further  evaluate our algorithm on more complicated \textsf{TUMKitchen}~\cite{tenorth09dataset} and \textsf{HDM05}~\cite{cg-2007-2} datasets: for various feature vectors applied, 
 our algorithm achieves better or similar precision/recall values but runs significantly faster than the state-of-the-art.

\section{Related Work}
\label{section:related_work}

Temporal action segmentation is an important research topic with applications in robotics, machine learning, computer vision, etc. 
The model-based approach for solving this problem relates to action detection, which is usually addressed by training discriminative 
or generative models~\cite{oneata2013action,lea2016temporal,li2016online,pirsiavash2014parsing,richard2016temporal,tang2012learning,zhao2017temporal,layher2017real}.
 These  methods require either a large amount of training data with annotations which introduces high computation cost, or  the collection of training sets and an offline learning step, e.g., \cite{cheng2014temporal,kuehne2016end}.
 In contrast, unsupervised clustering-based methods  parse actions into motion primitives only based on the processed data stream,
 and hence  can be used
 in many important scenarios in which training sets are not publicly available, e.g., health-care and surveillance.

\cite{ZhouDH13}  proposes aligned cluster analysis~(\textsf{ACA}) that uses a novel kernel for time series alignment. After specifying the number of clusters, as well as the minimum and maximum lengths of the action segments, both the  \textsf{ACA} and \textsf{HACA}, an improved version of \textsf{ACA}, solve a dynamic program over the entire stream, which makes both approaches non-applicable for fast and online video segmentation.
\cite{kruger2017efficient} proposes an efficient motion segmentation approach, in which  a novel feature bundling method is used to generate compact and robust motion representations. 
In particular, a generalised search radius is introduced in \cite{kruger2017efficient}  so that 
the number of clusters is not needed as input. 
\cite{li2015temporal} addresses motion segmentation via subspace clustering, in which a temporal Laplacian regularisation method is applied to encode the temporal structure in the subspace and spectral clustering is applied to group actions using the projected feature vectors. 

Another related line of research  is the mixture model for clustering, which can be trained online. 
\cite{verbeek2003efficient} proposes a greedy method for training the Gaussian mixture model,  but requires to specify 
the model complexity, e.g., the number of clusters $k$, in advance.  To overcome this, the Bayesian nonparametric models learn the value of $k$ from the input data jointly with the observation models.  For example,
Dirichlet process mixture model (\textsf{DPMM})~\cite{blei2006variational,gershman2012tutorial} uses a Dirichlet process to allocate samples and calculate $k$, where the joint probabilities of allocations are independent of the temporal order of the input samples.  \cite{hughes2013memoized}  proposes an algorithm to train \textsf{DPMM}  by iteratively processing data batches and allowing the number of clusters varying when  the model parameters are updated online.

Our work differs from all these previous methods. Most previous studies formulate the human action segmentation via some optimisation problem, and apply dynamic programming or spectral techniques to solve the problem. These techniques usually require the global information of the dataset and have high computational cost, in contrast to our greedy and online 
approach. Comparing our approach with \textsf{DPMM}, one should notice that \textsf{DPMM} is based on keeping track of some probabilistic distributions, while our algorithm keeps track of some geometric measure associated with the input.

Regarding the online setting,
\cite{gong2012kernelized} proposes Kernelised Temporal Cut~(\textsf{KTC}) to sequentially cut a video stream into different snippets, and develops an online algorithm to locate segment boundaries.  \cite{hasan2016learning}  
proposes an approach to learning the temporal regularity based on the autoencoder. Then the local minima of the regularity scores can be regarded as action transitions. Our work differs from these mentioned online methods: They perform segmentation based on changing point detection while our method is based on online clustering. Thus, our algorithm produces a \emph{clustering} of the segmented snippets, and the output of our algorithm can provide more semantic meaning than other online approaches.

\section{Algorithm}
\label{section:algorithm}

Now we present our dynamic clustering method for segmenting human actions, which  refers to the task of dividing a video stream into disjoint snippets so that every snippet presents a single action.  This problem is closely related to partitioning the associated feature vectors into clusters such that the feature vectors in each cluster represent a single human action. 
 
 Formally, we assume that we are given a sequence of vectors $\vecx_1,\ldots, \vecx_t,\ldots$ as input, in which every $\vecx_i\in\mathbb{R}^d$ is the feature vector at time $i$.  The goal is to find cut points~(the transition timestamps), such that different human actions are separated by these cut points. 
 To make our  approach applicable in realistic scenarios, we do not make additional assumptions on the input feature vectors, in the sense that each new arriving input $\vecx_i$ could be either a feature vector corresponding to a single frame or a video snippet~(e.g., a Fisher vector~\cite{kuehne2016end,Wang2013} can represent the context of $50$ frames).

 \subsection{Our Approach}
 
The design of our algorithm is based on the fact that the feature vectors representing the same human actions are similar to each other, although a ``quantitative'' measurement of this similarity usually depends on the nature of the feature vectors, including the dimensionality, the spatiotemporal scales, and how a feature vector is generated, etc. To deal with this fact,
our  algorithm first computes  the basic statistics of the input feature vectors, which is achieved by building a fully connected similarity graph $G$ from the initial feature vectors of a fixed-length sliding window and running a spectral clustering algorithm on $G$, where the number of clusters $k$ is determined by the graph spectra \cite{Luxburg07}.  Comparing with a classical spectral clustering algorithm that only gives a partition of the vertices of $G$, our algorithm computes the centers $\{\mathbf{c}_i\}_{i=1}^k$ of the clusters to measure \emph{how far clusters are separated from each other}, and  diagonal matrices $\{\mathbf{\Sigma}_i\}_{i=1}^k$, each of which is associated with one cluster and measures \emph{how concentrated 
 all the feature vectors within each cluster are around the center}. We define the cluster set $\mathcal{C}$ as the combination of $\{\mathbf{c}_i\}_{i=1}^k$ and $\{\mathbf{\Sigma}_i\}_{i=1}^k$. The length of the time window is dependent on a specific application, in order to   cover the representative statistics of the entire time series. 
 
Afterwards, our algorithm enters the updating phase and runs in an online fashion: for every arriving  vector $\vecx_t$, the algorithm computes the distance between $\vecx_t$ and its closest center of all the observed clusters $\vecc_1,\ldots, \vecc_k$, i.e.,
$
\mathsf{dist}(\vecx_t, \mathcal{C}) = \min_{1\leq i\leq k} \|\vecx_t - \vecc_i \|^2.
$
Based on  $\mathsf{dist}(\vecx_t, \mathcal{C}) $ and  matrices $\{\mathbf{\Sigma}_i\}$, the algorithm  decides whether to put $\vecx_t$ into its closest cluster, or generates a new cluster to host $\vecx_t$.   Notice that the runtime of this step is only proportional to the number of currently observed clusters and the dimension of the feature vectors,  which is independent of the length of the video stream.
Hence, the update time for every arriving feature vector can be accomplished in $O(1)$ for most applications.




\subsection{Initialisation Step}


The input of the initialisation step are the initial  $\ell$ feature vectors $\mathbf{x}_1,\ldots, \mathbf{x}_{\ell}$ from a video stream. The algorithm first constructs a fully connected graph   $G=(V,E)$ of $\ell$ vertices, denoted by $v_1,\ldots, v_{\ell}$, where each $v_i$ corresponds to a vector $\vecx_i$ and the weight of the edge between $v_i$ and $v_j$ is defined as
$w(v_i, v_j) = \mathrm{exp} \left( - \| \vecx_i - \vecx_j \|^2/\sigma \right)$
with a parameter $\sigma$. This graph $G$ is represented by the normalised Laplacian matrix, which is defined by 
$
\mathcal{L} = \mathbf{I} - \mathbf{D}^{-1/2} \mathbf{A} \mathbf{D}^{-1/2}$,
where $\mathbf{I}\in\R^{\ell\times \ell}$ is the identity matrix, $\mathbf{D}\in\mathbb{R}^{\ell\times \ell}$ is the diagonal matrix defined by 
$
\mathbf{D}_{i,i} = \sum_{j} w(v_i, v_j)$,
 and $\mathbf{A}\in\R^{\ell\times \ell}$ is the adjacency matrix of  $G$. We then compute the eigenvalues $\lambda_1\leq\ldots\leq\lambda_{\ell}$ of $\mathcal{L}$, and use the smallest value 
 $k$ with a large gap between $\lambda_{k+1}$ and $\lambda_k$ to determine the number of clusters. This method is widely used for determining the number of clusters in practice, see  \cite{PSZ} for theoretical explanation and  \cite{Luxburg07} for detailed discussion.

After computing the value of $k$, the algorithm runs  $k$-means  for the initial $\ell$ feature vectors $\mathbf{x}_1,\ldots, \mathbf{x}_{\ell}$.  In addition to the $k$ clusters produced by $k$-means, the following quantities are computed: (1) Centres $\{\vecc_i\}_{i=1}^k$ for these $k$ clusters; (2)  Covariance matrices $\left\{\mathbf{\Sigma}_{i} \right\}_{i=1}^k$, where 
 $
 \mathbf{\Sigma}_{i} \triangleq \mathbf{diag} \left( \left(\sigma^{(i)}_1\right)^2,\ldots, \left(\sigma^{(i)}_d\right)^2 \right) 
 $
 and $\sigma^{(i)}_j$ is defined as the $\ell_2$-distance of the $j$th coordinate between all the feature vectors in the $i$th cluster and their center; (3) Radiuses $\{r_i\}_{i=1}^k$ for the $k$ clusters and defined by 
 $
 r_i\triangleq  \Tr \mathbf{\Sigma}_i$.
 Moreover, we define a minimum area of the cluster $\delta_r$ with
 $ \delta_r\triangleq c\cdot {d}$,
which is similar to the search radius introduced in \cite{kruger2017efficient}. The constant $c$ is a hyper-parameter, which can be viewed as the uncertainty estimate of each dimension,  and is empirically determined depending on the use.

\subsection{Online Evolution Step}

The second step of our algorithm is the online clustering based on the statistical information of the currently observed clusters. Specifically,  for every arriving $\vecx_t$ the algorithm computes the distance between $\vecx_t$ and its closest center, i.e.,
$
\mathsf{dist}(\vecx_t, \mathcal{C}) = \min_{1\leq i\leq k} \|\vecx_t - \vecc_i \|^2.
$
Depending on such distance, the algorithm either creates a new cluster consisting of a single point $\vecx_t$, or adds $\vecx_t$ to its closest cluster and updates the corresponding $\vecc_i$ and $\mathbf{\Sigma}_i$. The formal description is presented in Algorithm~\ref{update}. Here, $\circ$ denotes the Hadamard product operation. Notice that, for the task of online segmentation, a transitional point is detected if the current observed $\vecx_t$ and $\vecx_{t-1}$ belong to different clusters.



\begin{algorithm}
\caption{Online Evolution Step\label{update}}
\begin{algorithmic}[1]

\State $
\mathsf{dist}(\vecx_t, \mathcal{C}) = \min_{1\leq i\leq k} \|\vecx_t - \vecc_i \|^2.$

$i^{\star} = \argmin_{1\leq i\leq k} \|\vecx_t - \vecc_i \|^2.$

\If{$\mathsf{dist}(\vecx_t, \mathcal{C})\geq  \max\left\{ r_{i^{\star}}, \delta_r\right\}    $ } 

 $k \leftarrow k+1$; 
  $S_k\leftarrow\{ \vecx_t\}$; 
  
 $\mathbf{c}_k\leftarrow\vecx_t$;
 $M_{k} \leftarrow \vecx_t \circ  \vecx_t$;
   
 $\Sigma_{k} \leftarrow \mathbf{diag}\left( 0,0,...,0\right)$.

 \Else 

$\mathbf{c}_{i^{\star}} \leftarrow \mathbf{c}_{i^{\star}} + (\mathbf{x}_t - \mathbf{c}_{i^{\star}})\cdot \left|S_{i^{\star}} \right|^{-1}$;

$M_{i^{\star}} \leftarrow  \left(M_{i^{\star}} \cdot \left|S_{i^{\star}} \right| +  \vecx_t \circ  \vecx_t \right) / \left( \left|S_{i^{\star}} \right| + 1 \right)  
$;

$
\mathbf{\Sigma}_{i^*} \leftarrow M_{i^{\star}} - \mathbf{c}_{i^{\star}} \circ \mathbf{c}_{i^{\star}}
$;

 $
 r_{i^{\star}} \leftarrow  \Tr\mathbf{\Sigma}_{i^{\star}}
 $;
$S_{i^{\star}} \leftarrow S_{i^{\star}} + \{\vecx_t \}$.
 
 \EndIf
\end{algorithmic}
\end{algorithm}

\subsection{Runtime Analysis}

For the initialisation step,  we first need to build the similarity graph based on $\ell$ feature vectors, each of which has dimension $d$. This takes  $O(\ell^2\cdot d)$ time. After that, computing all the eigenvalues of matrix $\mathcal{L}$ takes 
$O(\ell^3)$ time,  and $O(\ell\cdot k) $ time is needed for writing down the spectral embedding of all the vertices. After this, we can apply a polynomial-time approximation scheme (PTAS) algorithm for the $k$-means clustering problem with runtime $O\left(\ell\cdot k^2+ 2^{\tilde{O}(k)}\right)$ when treating
the approximation parameter $\varepsilon$ involved in the definition of PTAS as $\varepsilon=\Omega(1)$, \cite{FeldmanMS07}. Since the number of clusters $k=O(1)$ for most applications, 
 this step takes  $O\left(\ell^2\cdot d + \ell^3  \right)$ time. 
 

For the online evolution step, computing the distance between every arriving $\vecx_t$ and its closest center takes time $O(k\cdot d)$, where $k$ is the number of clusters at time $t$. After that, $O(d)$ time suffices to update all the quantities maintained by the algorithm. Hence, the update time for each arriving feature vector is $O(k\cdot d)$.

\subsection{Further Discussion}

At the end, we briefly discuss our algorithm in comparison to online $k$-means clustering, which has
received considerable attention in  theoretical computer science and machine learning. It has been observed in \cite{LSS16} that any online $k$-means algorithm with bounded approximation guarantee has to generate strictly more than $k$ clusters, due to the following fact: 

Assume all the data points are in $\mathbb{R}$, $k=2$, and the first two arriving points are $v_1=0$ and $v_2=1$, respectively. At this point any algorithm will put $v_1$ and $v_2$ into two clusters with total cost $0$, since otherwise putting them into a single cluster will increase the cost value from $0$ to $1/2$.  However, after an algorithm assigns $v_1$ and $v_2$ to different clusters, the third arriving point could be $v_3=c$ for an arbitrary large value $c$. At this stage, the algorithm is forced to put $v_3$ into one of the two existing clusters, which makes the total cost as  least $\Omega(c)$ in contrast to the optimal cost value $1/2$. It is straightforward to generalise this example to the case of $k\geq 3$ clusters in high-dimensional space, which explains why online $k$-means algorithms cannot be applied to solve our problem.

However, our key observation is that, 
a combination of  determining the value of $k$ via graph spectra and a variant of $k$-means could work well for human action detection. To explain the reason, we look at the task of online clustering  points $\vecx_1,\ldots, \vecx_n\in\mathbb{R}^d$ into $k$ clusters with the following two extreme cases, where these $k$ clusters are  far  from each other. (i) When the points associated with each cluster arrive together, it's not difficult to find a nearly-optimal clustering; (ii) 
the task becomes challenging when the $k$ points from $k$ different clusters arrive first, since whether grouping them into $k$ or fewer clusters  depends on the locations of the remaining $n-k$ points, i.e., the worst-case  example for online $k$-means discussed above.
Remarkably, for online human action detection we do not  need to consider (ii) since  graph spectra  allow us to learn the geometry of the initial data points, e.g., how they're concentrated around the centers. Moreover, due to the physical continuity of human motions, the temporal structure of a feature sequence is smooth and an algorithm will   receive many points around a center, before receiving points from another cluster, and the statistical information of the initial clusters  can be usually generalised to new clusters. Such temporal coherence characteristic is also considered in other tasks, e.g.~\cite{wiskott2002slow,jayaraman2016slow}.
Hence, it suffices to consider the scenario similar to (i). This explains  why our algorithm works in theory, and the worst-case scenario for online $k$-means is  successfully avoided through an application of the eigengap heuristic~\cite{Luxburg07}.

\section{Offline  Action Segmentation} 
\label{section:action_segmentation}

We  experimentally compare our algorithm with the state-of-the-art when various features are applied as input. Our experiments are conducted on a computer equipped with Intel Core i7-6700K $4\text{GHz}$ $8$-core processors, the Geforce GTX 1080 graphics card, and 32GB RAM. The operating system is Ubuntu 16.04. Our proposed algorithm is implemented with C++ and compiled in Matlab R2017a. The source code of our algorithm will be published online after the double-blind review process.


\subsection{Datasets}
Our algorithm will be evaluated on the following three datasets:

\begin{itemize} 
\item The \textsf{CMUMAD} dataset~\cite{huang2014sequential}   contains 40 recordings from 20 subjects,  and  covers comprehensive body action types from three modalities: RGB video and 3D depth map with spatial resolution of $240 \times 324$ of pixels, and the 3D body skeleton with 20 joints per frame.  Each recording is about 2 -- 4 minutes, and consists of 35 actions which are performed continuously. A null action (standing) is performed between consecutive actions. 

\item The \textsf{TUMKitchen} dataset~\cite{tenorth09dataset} consists of $20$ recordings from multiple modalities. The camera network captures the scene from four viewpoints with $25$ fps, and every RGB frame is of the resolution $384 \times 288$ by pixels. The action labels are frame-wise, and provided for the left arm, the right arm and the torso separately.

\item The \textsf{HDM05} dataset~\cite{cg-2007-2} is a large motion capture dataset containing more than $3$ hours of human motion recordings. In our experiments we use the third category, i.e., \emph{Sports}, which is significantly more challenging than the other categories. The {\em Sport} category incorporates motion recordings from $11$ sports classes, each of which  contains approximate $50000-100000$ frames.


\end{itemize}




\subsection{Feature Extraction} 
\label{featureextraction}

We use five types of features from different modalities, aiming to represent actions from a comprehensive set of information sources. The first two are context-based features, and the other three are pose-based features: 

\begin{itemize}

{\item \textsf{IDT+FV}:} In our experiment the motion boundary histogram is extracted around the dense trajectories of all scales \cite{Wang2013}. The Gaussian mixture model with $32$ components is trained using the other video performed by the same person, and vice versa. If the amount of motion boundary histograms is larger than $10^6$, only $10^6$ of them are selected randomly to train the Gaussian mixture model. Instead of encoding an entire video, we use a sliding window scheme to aggregate temporally local features. The time window spans $50$ frames and the stride is $1$. The resulting Fisher vector has $12,288$ dimensions.

{\item \textsf{VGG16}:} We  use the two stream deep neural model VGG-16 \cite{simonyan2014two,Feichtenhofer16} which is pre-trained on the UCF101-Split1 dataset \cite{soomro2012ucf101}, and extract the output of the last fully-connected layer. Since the colour channel and the motion channel are separate, we concatenate the outputs from both channels to compose a feature of $202$ dimensions for our experiments. Due to the model architecture, the \textsf{VGG16} feature describes the context in a snippet of $10$ frames.

{\item \textsf{JointLocation}:}  We concatenate all the $3$D coordinates of the joints to form a feature of each individual frame. Such feature can be viewed as an unary representation of a body configuration.

{\item \textsf{RelativeAngle}:} We use the relative $2$D angle between two adjacent body components to represent the rotation. This feature can be viewed as a high-order representation of the body configuration and possesses rotational invariance. 

{\item \textsf{Quaternions}:} Like in \cite{gong2012kernelized}, we use the quaternions to represent the $3$D rotation of a joint. Due to the ability of representing $3$D rotations, the quaternion feature provides richer information than \textsf{RelativeAngle}.
\end{itemize}


\paragraph{Feature Aggregation.} Since features that only encode  information of one or a few frames are not sufficient to encode actions,  we apply feature aggregation  to derive action patterns from temporally local features. Feature aggregation consists of two step: feature encoding, and temporal pooling. For feature enconding,
we view the output of our algorithm as codebook and encode each individual local feature using soft assignment~\cite{peng2016bag}. To perform temporal pooling, we define several temporal windows, fuse (add and normalize) the encoded features within each temporal window and obtain the action patterns. Afterwards, action segmentation is obtained by simply performing $k$-means on such action patterns.

We use different temporal pooling approaches for different datasets. In each recording {\sf CMUMAD},  the null action comes first and repeatedly occurs between each two actions. We assign the label 0 to  the frames that belong to the  same cluster with the first frame. Afterwards, we apply Gaussian smoothing on the temporal label sequence and detect all the peaks as well as their widths to create time windows. In contrast, as the {\sf TUMKitchen} and {\sf HDM05} datasets do not have 
null actions,   we use a sliding window scheme to perform temporal pooling, in which each time window consists of  30 frames and the stride is of 1 frame.

\subsection{Results on the {\sf CMUMAD} Dataset}
\label{experiment_segmentation}
We first introduce the experimental setup and the evaluation metrics. Then we present our experimental results and compare our algorithm with the state-of-the-art methods.


\paragraph{Evaluation Metric.} The performance of all algorithms are evaluated based on precision-recall values~\cite{dreherframework}, in which the true positive is commonly defined based on a region matching without considering the labels~\cite{kruger2017efficient,li2015temporal}. Here we use a more critical definition to capture the semi-semantic meaning of the segmentations given by the clusters. We define a true positive as the segment within which the majority of the predicted labels have more than $50\%$ overlaps with the ground truth labels, and this predicted label is not $0$ and not predicated before. Hence the influence of null actions is excluded and each meaningful action can be regarded as a novel action. Moreover, we use runtime  to measure the speed of segmentation algorithms, for which the feature extraction step is not considered. All the reported values are the average of the ones from all the recordings.

\paragraph{Analysis with snippet-wise features.} We compare the performance of our algorithm with that of the classical $k$-means for snippet-wise features, without using the feature aggregation step for post processing. 
As shown in   Figure~\ref{compIV},   our algorithm runs much  faster than $k$-means, and achieves better precision and recall values. 
Our experiments also indicate that both of our algorithm and $k$-means perform better when applying the \textsf{IDT+FV} 
features rather than \textsf{VGG16} features: this is due to the fact that, while an \textsf{IDT+FV}  feature presents a snippet of $50$ frames, 
a \textsf{VGG16} feature  presents a snippet of $10$ frames and  is not sufficient to represent the actions spanning longer time.  

\begin{center}
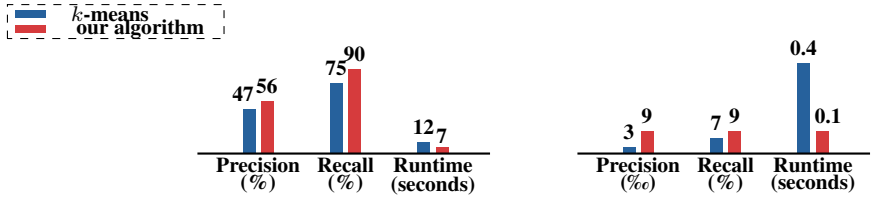
\begin{figure}[t]
\centering
\begin{tikzpicture}[scale=0.5]

\fill[mblue] (0,0) rectangle (0.32,1.175);

\node at (0,1.575) {\textbf{\scriptsize{47}}};

\fill[jasper] (0.48,0) rectangle (0.8,1.4);

\node at (0.63, 1.8) {\textbf{\scriptsize{56}}};

\node at (0.4,-0.3) {\textbf{\scriptsize{Precision}}};
\node at (0.4,-0.8) {\textbf{\scriptsize{(\%)}}};

\fill[mblue] (2.3,0) rectangle (2.62,1.875);
\fill[jasper] (2.78,0) rectangle (3.1,2.25);

\node at (2.45,2.275) {\textbf{\scriptsize{75}}};

\node at (2.93,2.65) {\textbf{\scriptsize{90}}};

\node at (2.7,-0.3) {\textbf{\scriptsize{Recall}}};

\node at (2.7,-0.8) {\textbf{\scriptsize{(\%)}}};

\fill[mblue] (4.6,0) rectangle (4.92,0.3);
\node at (4.75,0.7) {\textbf{\scriptsize{12}}};

\fill[jasper] (5.08,0) rectangle (5.4,0.175);

\node at (5.23,0.475) {\textbf{\scriptsize{7}}};

\node at (5,-0.3) {\textbf{\scriptsize{Runtime}}};

\node at (5,-0.85) {\textbf{\scriptsize{(seconds)}}};

\draw[thick] (-1.2,0) to (6.5,0);

\begin{scope}[shift={(10,0)}]
\fill[mblue] (0,0) rectangle (0.32,0.18);

\node at (0.15,0.58) {\textbf{\scriptsize{3}}};

\fill[jasper] (0.48,0) rectangle (0.8,0.6);

\node at (0.63, 1) {\textbf{\scriptsize{9}}};

\node at (0.4,-0.3) {\textbf{\scriptsize{Precision}}};
\node at (0.4,-0.8) {\textbf{\scriptsize{(\textperthousand)}}};

\fill[mblue] (2.3,0) rectangle (2.62,0.42);
\fill[jasper] (2.78,0) rectangle (3.1,0.6);

\node at (2.45,0.82) {\textbf{\scriptsize{7}}};

\node at (2.93,1) {\textbf{\scriptsize{9}}};

\node at (2.7,-0.3) {\textbf{\scriptsize{Recall}}};

\node at (2.7,-0.8) {\textbf{\scriptsize{(\%)}}};

\fill[mblue] (4.6,0) rectangle (4.92,2.4);
\node at (4.75,2.8) {\textbf{\scriptsize{0.4}}};

\fill[jasper] (5.08,0) rectangle (5.4,0.6);

\node at (5.4,1) {\textbf{\scriptsize{0.1}}};

\node at (5,-0.3) {\textbf{\scriptsize{Runtime}}};

\node at (5,-0.8) {\textbf{\scriptsize{(seconds)}}};

\draw[thick] (-1.2,0) to (6.5,0);
\end{scope}

\begin{scope}[shift= {(-11, 1.5)}]
\fill[mblue] (5,2.1) rectangle (5.8,2.3); 
\fill[jasper] (5,1.7) rectangle (5.8,1.9);

\node at (7.5, 2.2) {\textbf{\scriptsize{$k$-means}}};

\node at (8.3, 1.8) {\textbf{\scriptsize{our algorithm}}};

\draw[dashed] (4.8,1.55) rectangle (10.5,2.45);
\end{scope}

\end{tikzpicture}
\caption{Comparison between our algorithm and $k$-means when applying \textsf{IDT+FV} with 12,288 dimensions and \textsf{VGG16} with 202 dimensions as input. \label{compIV}}

\end{figure}
\end{center}

\vspace{-2em}

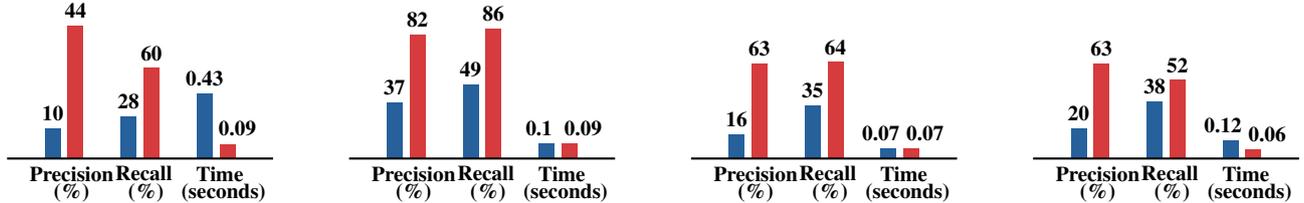
\begin{figure*}
\begin{center}

\begin{tikzpicture}


\begin{scope}[shift= {(0, 0)}]

\fill[mblue] (0,0) rectangle (0.2,0.4); \node at (0.1,0.6) {\textbf{\scriptsize{10}}};
\fill[jasper] (0.3,0) rectangle (0.5,1.76); \node at (0.4, 1.96) {\textbf{\scriptsize{44}}};

\node at (0.35,-0.2) {\textbf{\scriptsize{Precision}}}; \node at (0.35,-0.45) {\textbf{\scriptsize{(\%)}}};
\end{scope}

\begin{scope}[shift= {(1, 0)}]

\fill[mblue] (0,0) rectangle (0.2,0.56); \node at (0.1,0.76) {\textbf{\scriptsize{28}}};
\fill[jasper] (0.3,0) rectangle (0.5,1.2); \node at (0.4, 1.4) {\textbf{\scriptsize{60}}};

\node at (0.3,-0.2) {\textbf{\scriptsize{Recall}}}; \node at (0.33,-0.45) {\textbf{\scriptsize{(\%)}}};
\end{scope}

\begin{scope}[shift= {(2, 0)}]

\fill[mblue] (0,0) rectangle (0.2,0.86); \node at (0.1,1.06) {\textbf{\scriptsize{0.43}}};
\fill[jasper] (0.3,0) rectangle (0.5,0.18); \node at (0.53, 0.4) {\textbf{\scriptsize{0.09}}};

\node at (0.3,-0.2) {\textbf{\scriptsize{Time}}}; \node at (0.35,-0.45) {\textbf{\scriptsize{(seconds)}}};
\end{scope}

\draw[thick] (-0.5,0) to (3,0);


\begin{scope}[shift= {(4.5, 0)}]

\fill[mblue] (0,0) rectangle (0.2,0.74); \node at (0.1,0.94) {\textbf{\scriptsize{37}}};
\fill[jasper] (0.3,0) rectangle (0.5,1.64); \node at (0.4, 1.84) {\textbf{\scriptsize{82}}};

\node at (0.35,-0.2) {\textbf{\scriptsize{Precision}}}; \node at (0.35,-0.45) {\textbf{\scriptsize{(\%)}}};
\end{scope}

\begin{scope}[shift= {(5.5, 0)}]

\fill[mblue] (0,0) rectangle (0.2,0.98); \node at (0.1,1.18) {\textbf{\scriptsize{49}}};
\fill[jasper] (0.3,0) rectangle (0.5,1.72); \node at (0.4, 1.92) {\textbf{\scriptsize{86}}};

\node at (0.3,-0.2) {\textbf{\scriptsize{Recall}}}; \node at (0.33,-0.45) {\textbf{\scriptsize{(\%)}}};
\end{scope}

\begin{scope}[shift= {(6.5, 0)}]

\fill[mblue] (0,0) rectangle (0.2,0.2); \node at (0.0,0.4) {\textbf{\scriptsize{0.1}}};
\fill[jasper] (0.3,0) rectangle (0.5,0.2); \node at (0.58, 0.4) {\textbf{\scriptsize{0.09}}};

\node at (0.3,-0.2) {\textbf{\scriptsize{Time}}}; \node at (0.35,-0.45) {\textbf{\scriptsize{(seconds)}}};
\end{scope}

\draw[thick] (4,0) to (7.5,0);


\begin{scope}[shift= {(9, 0)}]

\fill[mblue] (0,0) rectangle (0.2,0.32); \node at (0.1,0.52) {\textbf{\scriptsize{16}}};
\fill[jasper] (0.3,0) rectangle (0.5,1.26); \node at (0.4, 1.46) {\textbf{\scriptsize{63}}};

\node at (0.35,-0.2) {\textbf{\scriptsize{Precision}}}; \node at (0.35,-0.45) {\textbf{\scriptsize{(\%)}}};
\end{scope}

\begin{scope}[shift= {(10, 0)}]

\fill[mblue] (0,0) rectangle (0.2,0.7); \node at (0.1,0.9) {\textbf{\scriptsize{35}}};
\fill[jasper] (0.3,0) rectangle (0.5,1.28); \node at (0.4, 1.48) {\textbf{\scriptsize{64}}};

\node at (0.3,-0.2) {\textbf{\scriptsize{Recall}}}; \node at (0.33,-0.45) {\textbf{\scriptsize{(\%)}}};
\end{scope}

\begin{scope}[shift= {(11, 0)}]

\fill[mblue] (0,0) rectangle (0.2,0.14); \node at (0.0,0.34) {\textbf{\scriptsize{0.07}}};
\fill[jasper] (0.3,0) rectangle (0.5,0.14); \node at (0.58, 0.34) {\textbf{\scriptsize{0.07}}};

\node at (0.3,-0.2) {\textbf{\scriptsize{Time}}}; \node at (0.35,-0.45) {\textbf{\scriptsize{(seconds)}}};
\end{scope}

\draw[thick] (8.5,0) to (12,0);


\begin{scope}[shift= {(13.5, 0)}]

\fill[mblue] (0,0) rectangle (0.2,0.4); \node at (0.1,0.6) {\textbf{\scriptsize{20}}};
\fill[jasper] (0.3,0) rectangle (0.5,1.26); \node at (0.4, 1.46) {\textbf{\scriptsize{63}}};

\node at (0.35,-0.2) {\textbf{\scriptsize{Precision}}}; \node at (0.35,-0.45) {\textbf{\scriptsize{(\%)}}};
\end{scope}

\begin{scope}[shift= {(14.5, 0)}]

\fill[mblue] (0,0) rectangle (0.2,0.76); \node at (0.1,0.96) {\textbf{\scriptsize{38}}};
\fill[jasper] (0.3,0) rectangle (0.5,1.04); \node at (0.4, 1.24) {\textbf{\scriptsize{52}}};

\node at (0.3,-0.2) {\textbf{\scriptsize{Recall}}}; \node at (0.33,-0.45) {\textbf{\scriptsize{(\%)}}};
\end{scope}

\begin{scope}[shift= {(15.5, 0)}]

\fill[mblue] (0,0) rectangle (0.2,0.24); \node at (0.0,0.44) {\textbf{\scriptsize{0.12}}};
\fill[jasper] (0.3,0) rectangle (0.5,0.12); \node at (0.58, 0.32) {\textbf{\scriptsize{0.06}}};

\node at (0.3,-0.2) {\textbf{\scriptsize{Time}}}; \node at (0.35,-0.45) {\textbf{\scriptsize{(seconds)}}};
\end{scope}

\draw[thick] (13,0) to (16.5,0);

\end{tikzpicture}

\end{center}
   \caption{Comparison between our algorithm (red) with $k$-means (blue). Here, the four subfigures represent the comparison results when VGC16~(202 dimensions), JointLocation~(60 dimensions), RelativeAngle~(15 dimensions), and Quaternions~(60  dimensions) are applied as feature vectors. \label{result2}}
\label{fig:short}
\end{figure*}

 \paragraph{Analysis with frame-wise features.} We apply our clustering algorithm together with the feature aggregation step as post processing when frame-wise features are applied. Due to the experimental results before, we  apply \textsf{VGG16} features  as input for our experiments here. The evaluation result is shown in 
 Figure~\ref{result2}, which confirms again that  
  with similar runtime our algorithm produces better  results than  $k$-means  for all the tested features. Moreover, one can note that our algorithm runs much faster than $k$-means  for high dimensional feature vectors.
   

\paragraph{Comparing with the state-of-the-art.} {We compare the performance of our algorithm with the following state-of-the-art methods:}
\begin{itemize}
{\item Spectral clustering~(\textsf{SC}).} 
We implement the standard spectral clustering algorithm, and set the parameter  $\sigma$ associated with the similarity graph to  $\sigma=1$. We also set $k=36$, since the initial null action needs to be treated as a single cluster. 

{\item Temporal subspace clustering~(\textsf{TSC})~\cite{li2015temporal}.} 
We use the default hyper-parameters specified in \cite{li2015temporal} for feature vectors of $100$ dimensions.  When the dimension of the input feature vectors is more than $100$, we first apply PCA to project all the input features to  $\mathbb{R}^{100}$.

{\item Aligned cluster analysis~(\textsf{ACA})~\cite{ZhouDH13}:}
We use the implementation of \textsf{ACA} provided by the authors of \cite{ZhouDH13}, in which the number of clusters is set to $36$ and the minimal and maximal segment lengths are tuned based on the CMUMAD dataset. We remark that we do not compare our algorithm with the hierarchical \textsf{ACA} algorithm (\textsf{HACA}), because of much higher computational cost and hyper-parameter tuning for the latter algorithm. Moreover, the \textsf{ACA} algorithm with the \textsf{IDT+FV} is computationally prohibitive. Thus, we first use PCA to project the Fisher vectors to $\mathbb{R}^{100}$ as well.

{\item Efficient Motion Segmentation~(\textsf{EMS})~\cite{kruger2017efficient}:}
We tune the hyper-parameters  to achieve the best performance with each feature. In addition, due to high computational load for \textsf{IDT+FV}, we first use PCA to project the Fisher vectors to a $100$-dimensional subspace. 
\footnote{The codes provided by the authors of \cite{kruger2017efficient} are implemented at Windows-64 operating system, 
and therefore this approach is tested on a computer equipped with Intel Core i7-6700HQ CPU at 2.6GHz and 16GB RAM.}

{\item Dirichlet Process Mixture Model~(\textsf{DPMM})~\cite{hughes2013memoized}.} 
We use the method proposed in~\cite{hughes2013memoized} to train the DPMM, and  set the allocation model as the Dirichlet process and the observation model as a Gaussian distribution with the diagonal covariance matrix. We denote \textsf{DPMM-A} as the combination of \textsf{DPMM} and our proposed feature aggregation method. 
\end{itemize}

\begin{figure}
\begin{center}
   \includegraphics[width=0.6\linewidth]{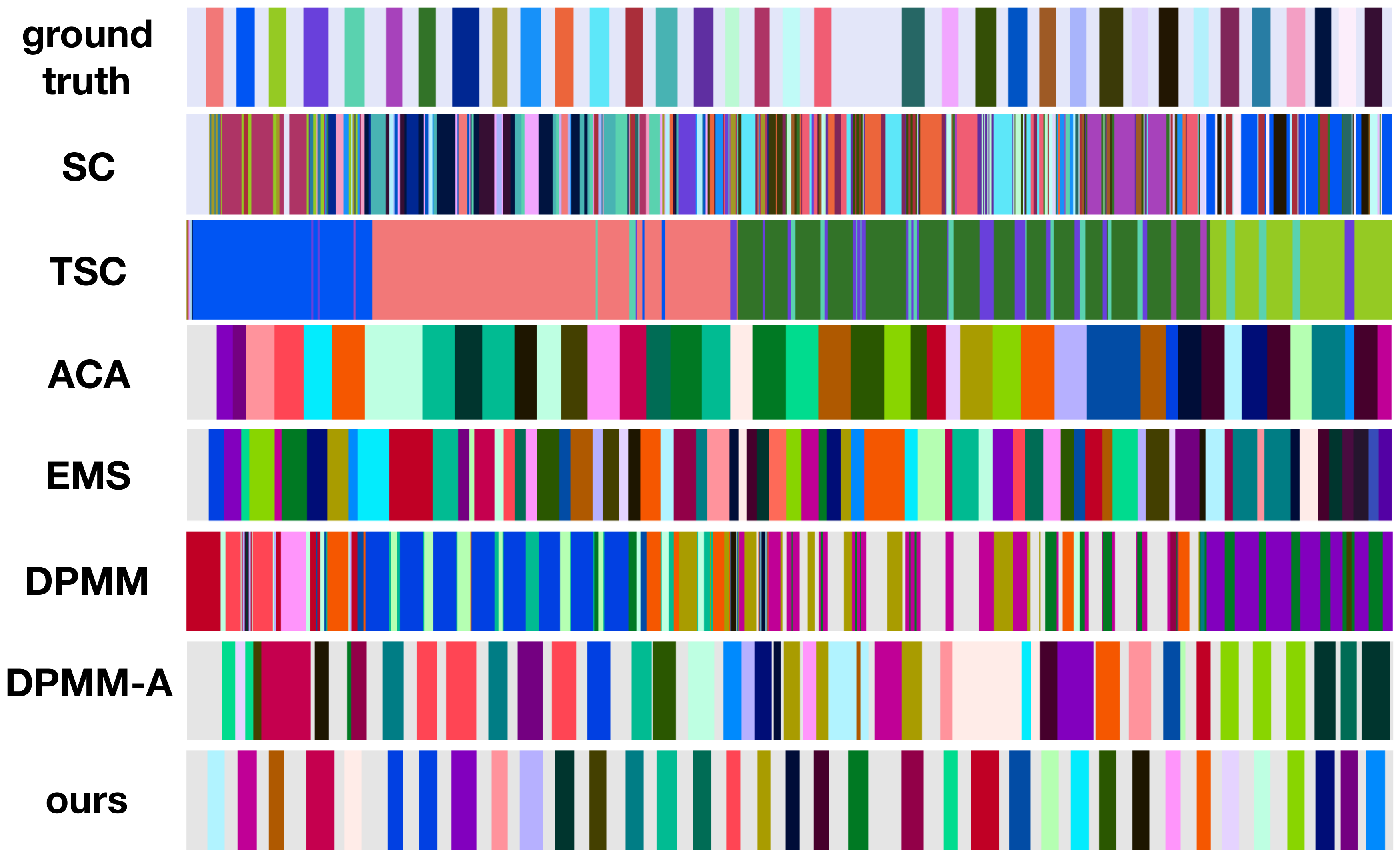}
\end{center}
   \caption{Visualised segmentation result of the compared algorithms. Here, the input is the first recording of Subject~5 of the dataset, and  \textsf{JointLocations} features are employed.   }
\label{fig:exp1_3}
\end{figure}

The comparison results are shown in Table~\ref{tab:exp1_3}:  \textsf{SC}, \textsf{TSC} and \textsf{DPMM}  are to assign each sample from a video stream to an {\em action cluster} directly, and hence perform better when grouping action patterns~ ({\bf IDT+FV}) than the other temporally local features, which lacks of representativeness of actions. Instead, \textsf{ACA} and \textsf{EMS} consider the similarities of feature vectors over a long period of  a video stream, which makes the output of  \textsf{ACA} and \textsf{EMS}
algorithms more consistent when different types of features are applied.  One can also clearly see that feature aggregation, which converts temporally local features to long-term segment patterns, boosts the performance of action segmentation when the temporally local features are applied as input, as shown in the results of \textsf{DPMM} and \textsf{DPMM-A}. When comparing our algorithm with the state-of-the-art, one can see that 
our algorithm runs hundreds of times faster than other algorithms but produces comparable results with \textsf{ACA}. In particular, when the   \textsf{JointLocation} features are applied, our algorithm produces the best results among all the tested algorithms, which is illustrated in Figure~\ref{fig:exp1_3}.

\begin{table}[t]
\small
\caption{
 Comparison between our method with state-of-the-art, where the results are given in  format of {\em precision/recall/runtime}. The best results are highlighted in boldface, and the time needed for PCA is included in the reported runtime. Since the feature {\bf IDT+FV} represents the content of 50 frames, feature aggregation is not applied \label{tab:exp1_3}}

\centering
\begin{tabular}{llllll}
\hline\noalign{\smallskip}
{ Algorithm} &{ \textsf{IDT+FV}}~\cite{Wang2013}&{ \textsf{VGG16}}~\cite{Feichtenhofer16} &{ \textsf{JointLocation}} & { \textsf{RelativeAngle}} &{ \textsf{Quaternions}} \\
\hline\noalign{\smallskip}

\multirow{1}{*}{\sf {SC}}  & {0.57/0.85/203.3} & {0.004/0.05/118.8} & {0.02/0.13/113.0}  & {0.003/0.06/113.4}  & {0.01/0.11/125.3} \\

\multirow{1}{*}{\sf {TSC}~\cite{li2015temporal}}   & {0.63/0.82/132.4} &{0.01/0.2/38.2} & {0.1/0.3/48.5} &{0.05/0.29/41.6}&
{0.05/0.29/38.7} \\ 
                                  
\multirow{1}{*}{\sf {ACA}~\cite{ZhouDH13}}   & {{\bf 0.91}/0.83/547.7} & {0.56/0.66/99.0} & {0.55/0.68/221.5} &{0.51/0.65/136.2} &{0.55/{\bf 0.66}/168.8}\\ 
                                                       
\multirow{1}{*}{\sf {EMS}~\cite{kruger2017efficient}} & {0.44/0.75/78.4 } & {{\bf0.67}/{\bf 0.73}/35.8} & {0.34/0.78/33.0} &{0.47/{\bf 0.89}/17.3} &{0.6/0.51/9.2 } \\

\multirow{1}{*}{\sf {DPMM}~\cite{hughes2013memoized}} & {0.4/0.73/507.8 } & {0.009/0.08/8.6} & {0.02/0.12/17.8} &{0.02/0.1/13.4} &{0.02/0.11/11.6 } \\     
                                                          
\multirow{1}{*}{\sf {DPMM-A}} & {n/a} & {0.24/0.53/8.6} & {0.37/0.54/17.8} &{0.27/0.5/13.4} &{0.39/0.58/11.6 } \\

\hline
\multirow{1}{*}{\sf {ours}}   & {0.56/{\bf 0.9}/{\bf 7.0}} & {0.44/0.6/{\bf 0.1}} & {{\bf 0.82}/{\bf 0.86}/{\bf 0.1}} &{{\bf 0.63}/0.64/{\bf 0.1}} &{{\bf 0.63}/0.52/{\bf 0.1}} \\ 
                                              
\hline

\end{tabular}
\end{table}

%

%
%
%
%

\subsection{Results on the \textsf{TUMKitchen} and \textsf{HDM05} Datasets}
\label{experiment_segmentation3}

We report our experimental results on the \textsf{TUMKitchen} and \textsf{HDM05} datasets, which contain more realistic and complex movements. 

\paragraph{Evaluation metric.} The \textsf{TUMKitchen} and \textsf{HDM05} datasets have no null actions and the performed actions might occur more than once, hence the strict evaluation metric used in the \textsf{CMUMAD} dataset is not suitable here anymore. To cope with this, we redefine the true positive in the precision-recall metric: we say that a ground truth boundary is correctly detected if the detected boundary by an algorithm is within $\pm 7$ frames~(about $\pm 0.25$ second) of the ground truth boundary. The reported values are the average of the experimental results from all recordings. 
 
\paragraph{Comparing with the state-of-the-art.} For the \textsf{TUMKitchen} dataset, action labels are annotated by individual body parts instead of an entire body. Therefore, we only apply the pose-based features (\textsf{JointLocation}, \textsf{RelativeAngle} and \textsf{Quaternions}) and perform segmentation for individual body parts. Since the left arm and the right arm have symmetric dynamics, we evaluate the two arms jointly.  The detailed comparison results are shown in Table~\ref{tabnew1}, from which one can see that our algorithm runs much faster than the others but produces better/comparable results, which was confirmed by our experiments for the \textsf{CMUMAD} dataset.



\begin{table}[htb]
\label{tab:exp1}
\caption{
  Results on the  \textsf{TUMKitchen} dataset, where the results are shown in format of \emph{precision/recall/runtime} and the best results are shown in boldface \label{tabnew1}}
\small
\centering
\begin{tabular}{lllll}
\hline\noalign{\smallskip}
{Algorithm} & {Type} & {JointLocation} & {RelativeAngle} &  {Quaternions} \\
\noalign{\smallskip}
\hline
\noalign{\smallskip}

\multirow{2}{*}{\sf {SC}} & \footnotesize{Torso}  & \footnotesize{0.02/0.01/9.1} & \footnotesize{0.24/0.28/8.6} &  \footnotesize{0.3/0.45/8.7} \\ 
                                    & \footnotesize{Arms} &  \footnotesize{0.29/0.47/8.8} & \footnotesize{0.17/{\bf 0.82}/8.8} &\footnotesize{0.18/{\bf 0.76}/8.6} \\

\hline
\multirow{2}{*}{\sf{TSC}} &\footnotesize{Torso}  & \footnotesize{0.12/0.04/10.1} & \footnotesize{0.28/0.1/10.2} & \footnotesize{0.31/0.19/10.1} \\ 
                                    & \footnotesize{Arms} &  \footnotesize{0.42/0.28/10.2} & \footnotesize{0.29/0.38/10.3} & \footnotesize{0.28/0.56/10.2} \\ 

\hline
\multirow{2}{*}{\sf {ACA}} & \footnotesize{Torso}  & \footnotesize{0.19/0.01/30.5} & \footnotesize{ 0.3/0.02/15.6} & \footnotesize{{\bf 0.4}/0.03/24.6} \\ 
                                    & \footnotesize{Arms} &  \footnotesize{0.36/0.08/24.6} & \footnotesize{0.36/0.1/19.6} & \footnotesize{{\bf 0.34}/0.09/26.1} \\                                  

\hline
\multirow{2}{*}{\sf {EMS}} & \footnotesize{Torso}  & \footnotesize{0.13/0.06/4.6} & \footnotesize{0.28/0.15/35.4} & \footnotesize{0.37/0.12/11.5} \\ 
                                    & \footnotesize{Arms} &  \footnotesize{0.26/0.12/5.4} & \footnotesize{{\bf 0.38}/0.27/12.7} & \footnotesize{{\bf 0.34}/0.09/8.4} \\   
                                    
\hline
\multirow{2}{*}{\sf {DPMM}} & \footnotesize{Torso}  & \footnotesize{0.27/{\bf 0.66}/11.1} & \footnotesize{0.18/{\bf 0.63}/11.3} & \footnotesize{0.30/{\bf 0.55}/3.3} \\ 
                                    & \footnotesize{Arms} &  \footnotesize{0.33/{\bf 0.52}/10.2} & \footnotesize{0.12/0.54/2.0} & \footnotesize{0.22/0.27/4.9} \\

\hline
\multirow{2}{*}{\sf {DPMM-A}} & \footnotesize{Torso}  & \footnotesize{0.23/0.15/11.1} & \footnotesize{0.36/0.16/11.3} & \footnotesize{0.37/0.19/3.3} \\ 
                                    & \footnotesize{Arms} &  \footnotesize{0.31/0.41/10.2} & \footnotesize{0.23/0.58/2.0} & \footnotesize{0.24/0.46/4.9} \\

\hline
\multirow{2}{*}{\sf {ours}} & \footnotesize{Torso}  & \footnotesize{{\bf 0.46}/0.15/{\bf 1.0}} & \footnotesize{{\bf 0.34}/0.12/{\bf 0.3}} & \footnotesize{{\bf 0.4}/0.26/{\bf 1.3}} \\ 
                                    & \footnotesize{Arms} &  \footnotesize{{\bf 0.49}/0.3/{\bf 1.0}} & \footnotesize{0.27/0.64/{\bf 1.0}} &\footnotesize{0.33/0.68/{\bf 0.3}} \\                             
\hline

\end{tabular}
\end{table}

For the {\sf HDM05} dataset, we use the pose-based features as the {\sf HDM05} dataset only contains motion capture data. Our experimental results are shown in Table~\ref{tabnew2}, and confirm once more about the performance of our algorithm over the state-of-the-art.  We highlight that, on both {\sf TUMKitchen} and {\sf HDM05} datasets our algorithm yields much better recall values than \textsf{ACA}. A probable reason is that our algorithm compares each frame with previous ones, which leads to higher temporal resolution than the segment comparison in \textsf{ACA} and hence is able to locate the segment boundary more precisely.

\begin{table}[htb]
\label{tab:tabnew2}
\caption{
 Experimental results on  Part-Scene 3~(Sports) of  \sf{HDM05}\label{tabnew2}}
\centering
\begin{tabular}{llll}
\hline\noalign{\smallskip}
{\small Algorithm} &{\small JointLocation} & {\small RelativeAngle} &{\small Quaternions} \\
\noalign{\smallskip}
\hline
\noalign{\smallskip}

\multirow{1}{*}{\sf\footnotesize {SC}}  & \footnotesize{0.15/{\bf 0.36}/88.6} & \footnotesize{0.11/ 0.32/145.2} & \footnotesize{0.14/{\bf 0.37}/145.9} \\

\multirow{1}{*}{\sf\footnotesize{TSC}}   & \footnotesize{0.16/0.29/46.5} &\footnotesize{0.13/0.2/59.8} & \footnotesize{0.14/0.28/74.2} \\ 
                                  
\multirow{1}{*}{\sf \footnotesize{ACA}}   & \footnotesize{0.11/0.04/443.3} & \footnotesize{{\bf 0.17}/0.07/382.8} & \footnotesize{0.13/0.05/343.3} \\ 
                                                       
\multirow{1}{*}{\sf \footnotesize{EMS}} & \footnotesize{{\bf 0.17}/0.11/6.3} & \footnotesize{0.11/0.1/26.6} & \footnotesize{0.14/0.11/16.5} \\ 

\multirow{1}{*}{\sf \footnotesize{DPMM}} & \footnotesize{ 0.1/0.24/23.1} & \footnotesize{0.07/0.1/7.5} & \footnotesize{0.13/0.27/16.3}  \\

\multirow{1}{*}{\sf \footnotesize{DPMM-A}} & \footnotesize{0.11/0.16/23.1} & \footnotesize{0.07/0.15/7.5} & \footnotesize{{\bf 0.15}/0.23/16.3}  \\

\hline
\multirow{1}{*}{\sf \footnotesize{ours}}   & \footnotesize{ 0.15/ 0.23/{\bf 1.0}} & \footnotesize{0.11/{\bf 0.33}/{\bf 1.3}} & \footnotesize{{\bf 0.15}/0.27/{\bf 1.3}} \\ 
                                              
\hline

\end{tabular}
\end{table}

\section{Online Action Transition Detection}
\label{section:transition_detection}

In this section, we compare our method with the state-of-the-art online methods. In such action transition detection task, the algorithm locates a segment boundary at the current time, if the current frame belongs to a different cluster from the previous frame. Due to frame-by-frame processing in the online setting, the feature aggregation step is disabled.


\paragraph{Evaluation metric.} We use the same evaluation metric as in Sec.~\ref{experiment_segmentation3}. 
Since the task online is more time-sensitive, 
the temporal range to define the true positive is set to be  $\pm 2$ frames~($\pm 0.04$ second).


\paragraph{Dataset and feature extraction.} We use the \textsf{TUMKitchen} dataset for experiment. Comparing with \textsf{CMUMAD} and \textsf{HDM05}, it has more realistic movements, multi-modal recordings, and separate annotations on the torso and the arms. These properties are suitable for analysing complex movements of individual body parts from multiple information sources. Besides the three pose-based features, we also use the contextual information from the video recordings.

\paragraph{Comparing with the state-of-the-art.} We compare the performance of our algorithm with the following state-of-the-art online methods: 
\begin{itemize}

\item Sequential kernelized temporal cut~(\textsf{KTC-S})~\cite{gong2012kernelized}.  
The \textsf{KTC-S} is a method for segmenting human actions online. Here, we set every time window to be $25$ frames, and use the degenerated version\footnote{We did not implement the full version of the algorithm proposed in \cite{gong2012kernelized} due to lack of information provided in the paper.}.
 The hyper-parameters of \textsf{KTC-S} are selected experimentally in order to achieve the best performance.

{\item Autoencoder with temporal regularity (ATR)~\cite{hasan2016learning}.} In~\cite{hasan2016learning} the regions with high reconstruction errors (low regularity scores) are treated as abnormalities. Here we assume that human motions at action transitions violate the temporal regularity and extend the original offline local minima searching method to an online version, using the same sliding window method with \textsf{KTC-S}. Due to its generability, we directly use the provided model from \cite{hasan2016learning}.



\end{itemize}

\paragraph{Results and discussion.} From Table~\ref{tab:exp2}, one can see that  our algorithm produces comparable precision values but much higher recall values than other methods. We claim that in the online setting the recall is more important than precision for many applications (e.g., healthcare and surveillance) which require online and effectively detecting potential fast and abnormal actions.
It is also noticeable that all the tested online algorithms suffer from over-segmentation issues and hence cause low precision values.


%

\setlength{\tabcolsep}{4pt}
\begin{table}[t]
\label{tab:exp2}
\caption{
 Comparison between our algorithm and the state-of-the-art. The results are in format of {\em precision/recall} and the best values are highlighted in boldface. One notice that {\sf ATR} extracts features from videos rather than poses. We show the results under the pose-based features only for comparison \label{tab:exp2}}

\small
\centering
\begin{tabular}{lllll}
\hline\noalign{\smallskip}
Algorithm &  Type  & JointLocation & RelativeAngle &  Quaternions \\
\noalign{\smallskip}
\hline
\noalign{\smallskip}
\multirow{2}{*}{\sf {KTC-S}} & \footnotesize{Torso}& \footnotesize{{\bf 0.08}/0.36} & \footnotesize{{\bf 0.08}/0.31} &  \footnotesize{{\bf 0.10}/0.40} \\ 
                             & \footnotesize{Arms} & \footnotesize{{\bf 0.12}/0.38} & \footnotesize{{\bf 0.13}/0.39} &\footnotesize{{\bf 0.14}/0.42} \\

\hline
\multirow{2}{*}{\sf{ATR}} &\footnotesize{Torso}  & \footnotesize{0.06/0.29} & \footnotesize{0.06/0.29} & \footnotesize{0.06/0.29} \\ 
                          & \footnotesize{Arms} &  \footnotesize{0.10/0.30} & \footnotesize{0.10/0.30} & \footnotesize{0.10/0.30} \\ 
\hline
\multirow{2}{*}{\sf {ours}} & \footnotesize{Torso} & \footnotesize{0.07/{\bf 0.62}} & \footnotesize{{\bf 0.08}/{\bf 0.49}} & \footnotesize{0.08/{\bf 0.50}} \\ 
                            & \footnotesize{Arms}& \footnotesize{0.10/{\bf 0.39}} & \footnotesize{0.11/{\bf 0.41}} &\footnotesize{0.08/{\bf 0.46}} \\                             
\hline
\end{tabular}
\end{table}
\setlength{\tabcolsep}{1.4pt}

\section{Conclusions}
To cooperate with insistent need of processing data streams, we propose a dynamic algorithm for segmenting and clustering human actions. Comparing with previous methods, our algorithm is fast, generic for various features and also improves the state-of-the-art offline approaches when combining with a feature aggregation step. We experimentally evaluate our algorithm against the state-of-the-art methods on the standard datasets when different features are applied as input: for all the tested datasets, our algorithm produces better/comparable results  but runs much faster than the state-of-the-art. Moreover, in the important scenario of online human action segmentation, with comparable precision value our algorithm presents a significant improvement over the state-of-the-art with respect to recall value, which is more important for many applications.

Our work leaves many interesting problems for further research from both theoretical and applicational perspectives and we mention two questions here: (1) Theoretical studies for the  connection between human action segmentation and clustering.
Since clustering human actions requires to group  feature vectors representing the same  human action, it seems to be more challenging than setting segment boundaries. However, our proposed clustering algorithm achieves better results than  the state-of-the-art for the segmentation task. We believe that this direction worths further studies under different setups.
 (2) online abnormal detection of human behaviours. 
 The two questions are of wide interest not only for researchers working in computer vision and machine learning, but also for cognitive psychologists.



{\small
\bibliographystyle{ieee}
\bibliography{egbib}
}

\end{document}